%% file: main.tex
\documentclass[runningheads]{llncs}

\usepackage[year=2024,ID=6288]{eccv}

\usepackage{colortbl}
\definecolor{mygray}{gray}{.9}
\usepackage{multirow}
\usepackage{fontawesome}
\usepackage{wrapfig,lipsum,booktabs}

\usepackage[misc]{ifsym}

\usepackage[dvipsnames]{xcolor}

\usepackage{eccvabbrv}

\usepackage{graphicx}
\usepackage{booktabs}

\usepackage[accsupp]{axessibility}  

\usepackage[pagebackref,breaklinks,colorlinks,citecolor=eccvblue]{hyperref}

\usepackage{orcidlink}

\newcommand{\M}{\boldsymbol{M}}
\newcommand{\f}{\boldsymbol{f}}
\newcommand{\y}{\boldsymbol{y}}

\newcommand{\namelongm}[1]{Dynamic feature aggregation}
\newcommand{\nameshortm}[1]{DFA}

\newcommand{\namelonge}[1]{Dynamic encoder}
\newcommand{\nameshorte}[1]{DynE}

\newcommand{\namelongh}[1]{Dynamic TAD head}
\newcommand{\nameshorth}[1]{DyHead}

\newcommand{\nameshort}[1]{DyFADet}

\newcommand\blfootnote[1]{%
  \begingroup
  \renewcommand\thefootnote{}\footnote{#1}%
  \addtocounter{footnote}{-1}%
  \endgroup
}

\begin{document}

\title{DyFADet: Dynamic Feature Aggregation for Temporal Action Detection} 

\titlerunning{DyFADet for TAD.}

\author{Le Yang\inst{1}$^*$\textsuperscript{(\Letter)}\orcidlink{0000-0001-8379-4915} \and
Ziwei Zheng\inst{1}$^*$ \orcidlink{ 0009-0000-4896-3293}\and
Yizeng Han\inst{2}\orcidlink{0000-0001-5706-8784} \and
Hao Cheng\inst{3}    \and \\
Shiji Song\inst{4}\orcidlink{0000-0001-7361-9283} \and 
Gao Huang\inst{4}\orcidlink{0000-0002-7251-0988} \and
Fan Li\inst{1} }

\authorrunning{L.~Yang, Z.~Zheng et al.}

\institute{Xi'an Jiaotong University\\
\email{\{yangle15, lifan\}@xjtu.edu.cn}\blfootnote{$*$ Equal contribution. \Letter~Corresponding author.}, 
\email{ziwei.zheng@stu.xjtu.edu.cn}\and
Alibaba Group $\ \ \ $ \email{yizeng38@gmail.com}
 \and
HKUST(GZ) $\ \ \ $
\email{hcheng046@connect.hkust-gz.edu.cn} \and
Tsinghua University\\
\email{\{shijis, gaohuang\}@tsinghua.edu.cn}
}




\maketitle

\input{sec/0_abstract}    
\input{sec/1_intro}

\input{sec/2_related}
\input{sec/3_method}

\input{sec/4_exp}

\input{sec/5_conclusion}

%
%
\bibliographystyle{splncs04}
\bibliography{main}

\input{sec/sup1}

\end{document}

%% file: sec/0_abstract.tex
\begin{abstract}

Recent proposed neural network-based Temporal Action Detection  (TAD) models are inherently limited to extracting the discriminative representations and modeling action instances with various lengths from complex scenes by shared-weights detection heads. Inspired by the successes in dynamic neural networks, in this paper, we build a novel dynamic feature aggregation (DFA) module that can simultaneously adapt kernel weights and receptive fields at different timestamps. Based on DFA, the proposed dynamic encoder layer aggregates the temporal features within the action time ranges and guarantees the discriminability of the extracted representations. Moreover, using DFA helps to develop a Dynamic TAD head (DyHead), which adaptively aggregates the multi-scale features with adjusted parameters and learned receptive fields better to detect the action instances with diverse ranges from videos. With the proposed encoder layer and DyHead, a new dynamic TAD model, DyFADet, achieves promising performance on a series of challenging TAD benchmarks, including HACS-Segment, THUMOS14, ActivityNet-1.3, Epic-Kitchen~100, Ego4D-Moment QueriesV1.0, and FineAction. Code is released to \url{https://github.com/yangle15/DyFADet-pytorch}.




\keywords{Temporal action detection \and Dynamic network architectures \and Video understanding}

\end{abstract}

%% file: sec/1_intro.tex
\section{Introduction}
\label{sec:intro}


As a challenging and essential task within the field of video understanding, \textbf{T}emporal \textbf{A}ction \textbf{D}etection (TAD) has received widespread attention in recent years. The target of TAD is to simultaneously recognize action categories and localize action temporal boundaries from an untrimmed video. Various methods have been developed to address this task, which can be mainly divided into two categories: 1) Two-stage methods (such as~\cite{sridhar2021class,qing2021temporal}), which first learn to generate the class-agnostic action proposals and then conduct classification and boundary-refinement in proposal-level. 2) One-stage methods~\cite{zhang2022actionformer,lin2021learning,shi2023tridet,tang2023temporalmaxer} classify each frame as well as locate temporal boundaries in an end-to-end manner, achieving better performance and becoming a popular line of TAD research currently.

However, accurately detecting an action from an untrimmed video remains a challenging task. On the one hand, the spatial redundancy in the adjacent frames along with the static feature extraction strategy can result in poor discriminability of learned representations, hindering the detection performance~\cite{shi2023tridet,tang2023temporalmaxer}. On the other hand, the head inadaptation issue can happen in common TAD designs~\cite{zhang2022actionformer}, where a static shared-weight head is used to detect action instances with diverse temporal lengths, leading to sub-optimal performance. Therefore, it is necessary to find a solution that can simultaneously address these two key issues in modern TAD models.




\begin{figure}[t]
    \centering
    \includegraphics[width=\textwidth]{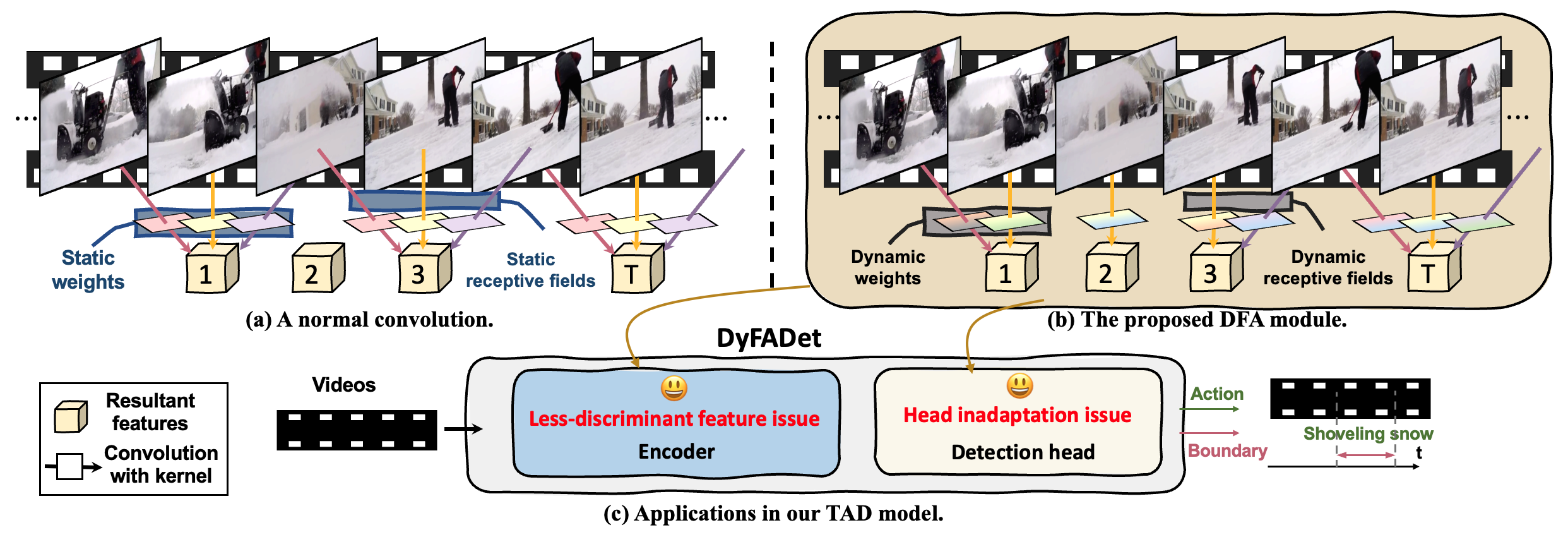}
    \vspace{-15pt}
    \caption{Differences between convolution and \nameshortm{}. (a) A normal convolution with static weights and receptive fields. (b) The dynamic formations of \nameshortm{} at different timestamps. (c) Implementing \nameshortm{} to build \nameshort{} can address the two issues in TAD.}
    \label{fig_1}
    \vspace{-20pt}
\end{figure}

In this paper, inspired by the recent success of dynamic neural networks~\cite{han2021dynamic,huang2023glance}, we develop a novel \textbf{D}ynamic \textbf{F}eature \textbf{A}ggregation (\nameshortm{}) module for TAD. As is shown in Fig.~\ref{fig_1}, the proposed \nameshortm{}, the kernel shape, and the network parameters can adapt based on input during generating new features, which significantly differs from the static feature learning procedure in most existing TAD models. Such a learning mechanism enables a dynamic feature aggregation procedure, which has the ability to increase the discriminability of the learned representations, and also adapts the feature learning procedure in detection heads based on each level of the feature pyramid during detection.

Therefore, based on the DFA, we build a dynamic encoder (\nameshorte{}) and a \namelongh{} (\nameshorth{}) to address the two issues in modern TAD models. On the one hand, the dynamic feature learning of \nameshorte{} can facilitate the features of target action to be gathered together and increase the differences between the features of the action and boundaries, resolving the first issue in Fig.~\ref{fig_1}. On the other hand, the \nameshorth{} will dynamically adjust detector parameters when it is applied at different pyramid levels, which corresponds to detecting the actions with different time ranges. By effectively addressing the two issues in Fig.~\ref{fig_1}, the proposed \nameshort{} with \nameshorte{} and \nameshorth{} can achieve accurate detection results in TAD tasks.

We evaluate the proposed \nameshort{} on a series of TAD datasets including HACS-Segment~\cite{zhao2019hacs}, THUMOS14~\cite{THUMOS14}, ActivityNet-1.3~\cite{caba2015activitynet}, Epic-Kitchen~100~\cite{damen2022rescaling}, Ego4D Moment Queries V1.0~\cite{grauman2022ego4d} and FineAction~\cite{liu2022fineaction}. The experimental results show the effectiveness of the proposed \nameshort{} in TAD tasks.

%% file: sec/2_related.tex
\section{Related Work}
\label{sec:related}

\subsubsection{Temporal action detection} is a challenging video understanding task, which involves localizing and classifying the actions in an untrimmed video. Conventional two-stage approaches \cite{qing2021temporal,sridhar2021class,xu2020g,zhu2021enriching,chen2022dcan,lin2019bmn,lin2018bsn,lin2021learning,yang2020revisiting,yang2023basictad} usually consist of two steps: proposal generation and classification. Nevertheless, these may suffer from high complexity and end-to-end training could be infeasible. A recent trend is designing one-stage frameworks and training the model in an end-to-end fashion. Some works \cite{liu2022end,shi2022react,tan2021relaxed} propose to detect actions with the DETR-like decoders \cite{carion2020end}, and another line of work \cite{cheng2022tallformer,zhang2022actionformer,shi2023tridet} builds a multi-scale feature pyramid followed by a detection head. In this work, we mainly follow the mainstream \emph{encoder-pyramid-head} framework. Concretely, we build our model based on the frameworks in the highly competitive methods \cite{zhang2022actionformer,shi2023tridet}, and boost the final detection performance by introducing the \emph{dynamic} mechanism to simultaneously solve the less-discriminant feature and the head inadaption issues in TAD models.

\vspace{-15pt}
\subsubsection{Dynamic neural networks} have attracted great research interests in recent years due to their favorable efficiency and representation power \cite{han2021dynamic}. Unlike conventional models that process different inputs with the same computation, dynamic networks can adapt their architectures and parameters conditioned on input samples~\cite{huang2018multi,yang2020resolution,han2022learning} or different spatial \cite{han2022latency,wang2021adaptive,han2024latency} / temporal \cite{meng2020ar,wu2019adaframe} positions. 

Data-dependent parameters have shown effectiveness in increasing the representation power with minor computational overhead \cite{yang2019condconv,ma2020weightnet,chen2020dynamic,dai2017deformable}. Existing approaches can generally be divided into two groups: one adopts mechanisms to dynamically re-weight parameter \emph{values} \cite{yang2019condconv,ma2020weightnet,chen2020dynamic}, including modern self-attention operators, which can be interpreted as (depth-wise) dynamic convolutions~\cite{zhou2023interpret}. While the static temporal reception fields of these methods can lead to the less-discriminant feature issue as stated in~\cite{shi2023tridet,tang2023temporalmaxer}. The other develops deformable convolution to achieve dynamic \emph{reception fields} \cite{dai2017deformable}, which have been effectively utilized in different video understanding tasks~\cite{lei2018temporal,mac2019learning,li2020spatio}. However, the kernel weights of these deformable convolutions are static. Compared to the previous methods, our \nameshortm{} \emph{simultaneously} adapts the convolution weights and temporal reception fields in a \emph{data-dependent} manner, leading to more effective and flexible modeling of temporal information in the TAD task. 


%% file: sec/3_method.tex
\section{Method}
\label{sec:method}

In this section, we first introduce the proposed \namelongm{} (\nameshortm{}) and then develop the dynamic feature learning based TAD model, \nameshort{} for TAD tasks. 

\subsection{\namelongm{}}

\begin{figure*}[htb]
    \centering
    \vspace{-15pt}
    \includegraphics[width=1.0\textwidth]{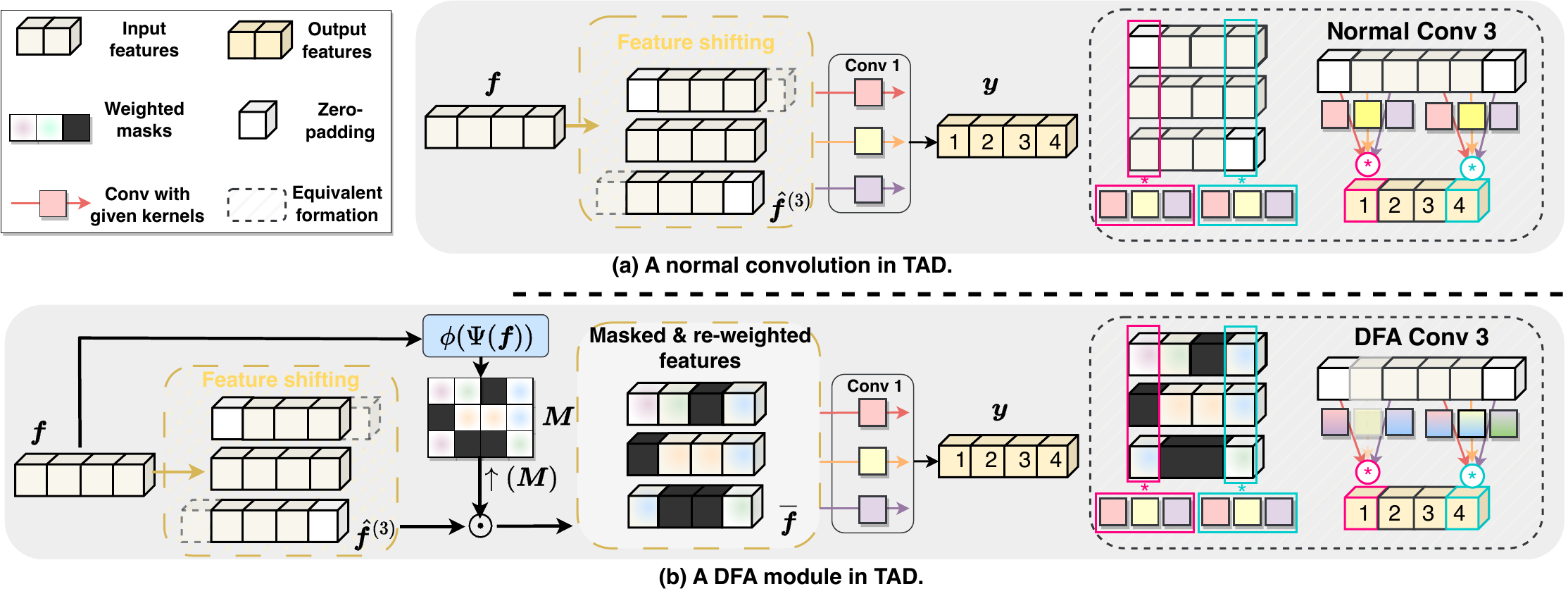}
    \vspace{-10pt}
    \caption{An illustration of the proposed \nameshortm{} module. (a) A normal convolution with the kernel size of 3 (Conv~3) and its corresponding formation realized by a shifting module and a point-wise convolution. (b) A \nameshortm{}. The shifted representations multiplied with the weighted mask will be sent to the point-wise convolution. The \nameshortm{} module is equivalent to a convolution with adaptive kernel weights and receptive fields.}
    \label{fig_dtfam}
    \vspace{-20pt}
\end{figure*}


\vspace{-10pt}
\subsubsection{DFA based on feature shifting.}
The \namelongm{} (\nameshortm{}) needs to effectively adapt the kernel weights and the receptive fields (including the number of sampling positions and the shape of kernels) based on the inputs to improve the feature extraction ability, which is difficult to simultaneously achieve. However, such a procedure can be realized if we separate a normal convolution into a feature-shifting module and a point-wise convolution~\cite{wu2018shift} (shown in Fig.~\ref{fig_dtfam}): A convolution with the kernel size of $k$ equals that the input features are first shifted according to the $k$ kernel positions, and then processed by a point-wise convolution (Conv~1). Motivated by this, we can learn an input-based mask, which will be used to zero and re-weight the shifted features. Then, using a Conv~1 to process the masked shifted features will equal a fully dynamic convolution shown in Fig~\ref{fig_dtfam} (b). A light-weight module, $\Psi(\cdot)$, and a non-linear activation, $\phi(\cdot)$ can be used to generate the mask, where the activation function can be the one whose output is always 0 with any negative inputs (such as ReLU~\cite{agarap2018deep} or a general-restricted tanh function~\cite{song2020fine}).

Formally, suppose we have a standard convolution with the kernel $\mathcal{K} \in \mathbb{R}^{C_{\text{out}}\times C_{\text{in}} \times k}$, where $k$ is the kernel size and $C_{\text{out}}$, $C_{\text{in}}$ are the numbers of input and output channels. Given an input $\f \in \mathbb{R}^{C_{\text{in}} \times T}$ ($T$ is the length of temporal dimension and $f_t \in \mathbb{R}^{C_{\text{in}}}$, is the tensor at time $t$), then we can shift the input by
\begin{small}    
\begin{align}
     &\hat{\f}^{(k)} = \text{Shift}(\f, k) = \left[ \hat{\f}^0 ;\  ...\  ;\hat{\f}^{k-1} \right] \in \mathbb{R}^{kC_{\text{in}} \times T} , \\ 
     &\hat{f}^{s}_t = f_{t-\lfloor k/2\rfloor+s}, \ s=0,1,...,k-1, \  t = 1,2,...,T,  \nonumber
\end{align}
\end{small}
where $\hat{\f}^s \in \mathbb{R}^{C_{\text{in}}\times T}$, the empty positions of the shifted features will be padded by all-zero tensors. The weighted masks can be calculated by 
\begin{small}
\begin{align}
    \label{eq2}
     \M = \phi(\ \Psi(\f)\ ) , \ \   \M \in \mathbb{R}^{C_{\text{m}} \times T},
\end{align}
\end{small}
where $C_{\text{m}}$ and $\Psi$ can be designed into different formations to achieve different dynamic properties. By repeated unpsampling the $\M$ to the dimension of $\hat{\f}^{(k)}$ (denoted by $\uparrow(\cdot)$), we have
\begin{small}
\begin{align}
     \overline{\f}\  =\  \uparrow(\M) \odot \hat{\f}^{(k)} = \overline{\M} \odot \hat{\f}^{(k)},
\end{align}
\end{small}
where we use $\overline{\M}$ to represent $\uparrow(\M)$ for simplicity, and $\odot$ means element-wise multiplication. The final output features can be written as
\begin{small}
\begin{align}
     \y = \text{\nameshortm{}} (\f) = \sum_{s=0}^{k-1} \mathcal{K}_s ( \overline{\M}_{[sC_{\text{in}}+1\, :\, (s+1)C_{\text{in}}]} \odot \hat{\f}^s ),
\end{align}
\end{small}
where $\mathcal{K}_s \in \mathbb{R}^{C_{\text{out}}\times C_{\text{in}}}$, $\overline{\M}_{[(sC_{\text{in}}+1\, :\, (s+1)C_{\text{in}}]} \in \mathbb{R}^{C_{\text{in}}\times T}$ means the mask tensor with the index from $sC_{\text{in}}+1$ to $(s+1)C_{\text{in}}$, and $\y \in \mathbb{R}^{C_{\text{out}} \times T}$ is the output.

\subsubsection{Different formations of \nameshortm{}.}
Using different $\Psi(\cdot)$ will result in the different formations of \nameshortm{}. By implementing $\Psi(\cdot)$ as a convolution, the \nameshortm{} can be built into a convolution-based \nameshortm{} module (DFA\_Conv). Moreover, changing the $C_{\text{m}}$ leads to the different dynamic properties. Take DFA\_Conv as an example. K-formation:  using $C_{\text{m}} = k$, the $\uparrow(\cdot)$ will repeat the mask in the interleaved manner at the channel dimensions with $C_{\text{in}}$ times, which makes the \nameshortm{} share the same dynamic receptive field among the channel dimension. C-formation: For $C_{\text{m}} = C_{\text{in}}$, the $\uparrow(\cdot)$ will repeat the mask $k$ times at the channel dimensions, which makes the \nameshortm{} a temporal dynamic channel pruning convolution. Also, using $C_{\text{m}} = kC_{\text{in}}$ (CK), the \nameshortm{} will adapt the receptive fields at different timestamps and channels.

\begin{figure}[htp]
\vspace{-10pt}
    \centering
    \includegraphics[width=0.95\linewidth]{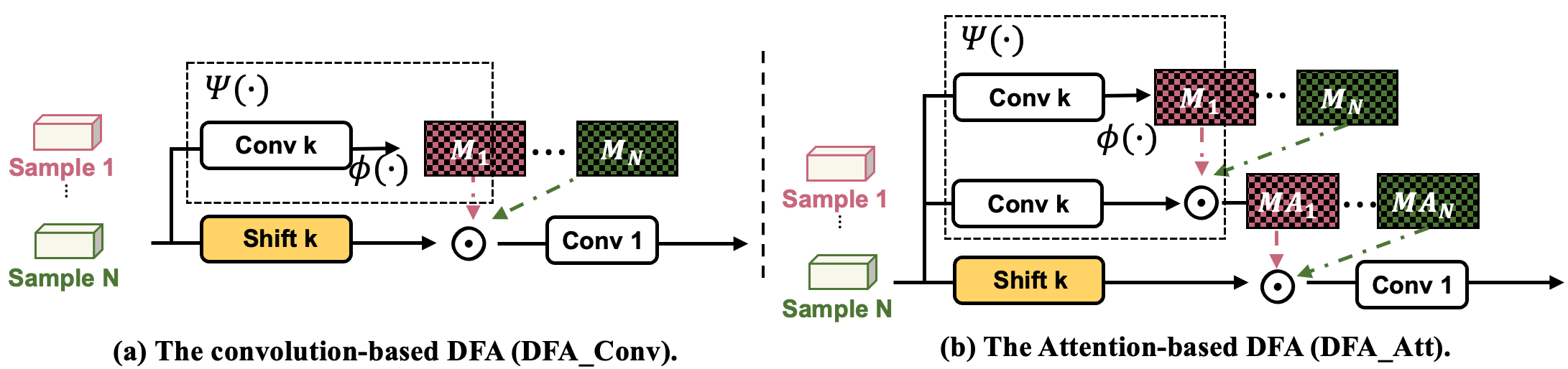}
    \caption{Two different formations of the proposed DFA module. }
    \label{fig_dye}
    \vspace{-20pt}
\end{figure}

Moreover, inspired by the success of Transformer-based architecture, we further implemented the \nameshortm{} in as the formation in Fig.~\ref{fig_dye} (b). The generated mask will result in the zeros in the attention matrix at the unimportant timestamps, and then the masked attentions will be used to re-weight the shifted features. 

\vspace{-10pt}
\subsubsection{Differences to existing works.}
If we restrict that $\M\in\{0,1\}$ has the same number of masked positions, then \nameshortm{} will equal to 1-d deformable convolution~\cite{dai2017deformable,lei2018temporal}, which adapts the shape of the kernels at different timestamp. If we remove the $\phi(\cdot)$, the \nameshortm{} will equal to the dynamic convolution in \cite{yang2019condconv,chen2020dynamic} which only adapts the kernel weights-based inputs. Moreover, TD3DConv in~\cite{li2020spatio} first uses temporal attention weights to adjust the features and then uses DC for feature learning, where these weights(W) are input-dependent yet temporal static during generating new features. While these weights are fully dynamic in our DFA. The \nameshortm{} provides an effective way for dynamic feature aggregation, which addresses the aforementioned issues in modern TAD models. Therefore, based on it, we further develop two important components \namelonge{} and \namelongh{} for the proposed TAD model, \nameshort{}.

\vspace{-10pt}
\subsection{\nameshort{}}

\subsubsection{Overview}
Based on the dynamic property of \nameshortm{}, we can build a TAD model that can effectively extract the discriminative representations in the encoder and adapt the detection heads for the action instances with different ranges. Following the common design in~\cite{zhang2022actionformer,shi2023tridet}, the proposed \nameshort{} consists of a video feature backbone, an encoder, and two detection heads for action classification and localization. Concretely, we first extract the video features using a pre-trained backbone. Then the extracted representations will be sent to the encoder to generate the features pyramid, where the former features will be down-sampled by the \nameshorte{} layer with the scale of 2 to obtain $\f^l, l =1,...,L$ ($L$ is the number of total levels). These features will be then used by \nameshorth{} for action detection. The architecture of \nameshort{} is shown in Fig.~\ref{fig_over} (a), where the \nameshorte{} layer and \nameshorth{} are two components built based on \nameshortm{} applying the dynamic feature learning strategy.

\subsubsection{Feature encoder with \nameshorte{}.}
\begin{figure*}[t!]
    \centering
    \includegraphics[width=0.95\textwidth]{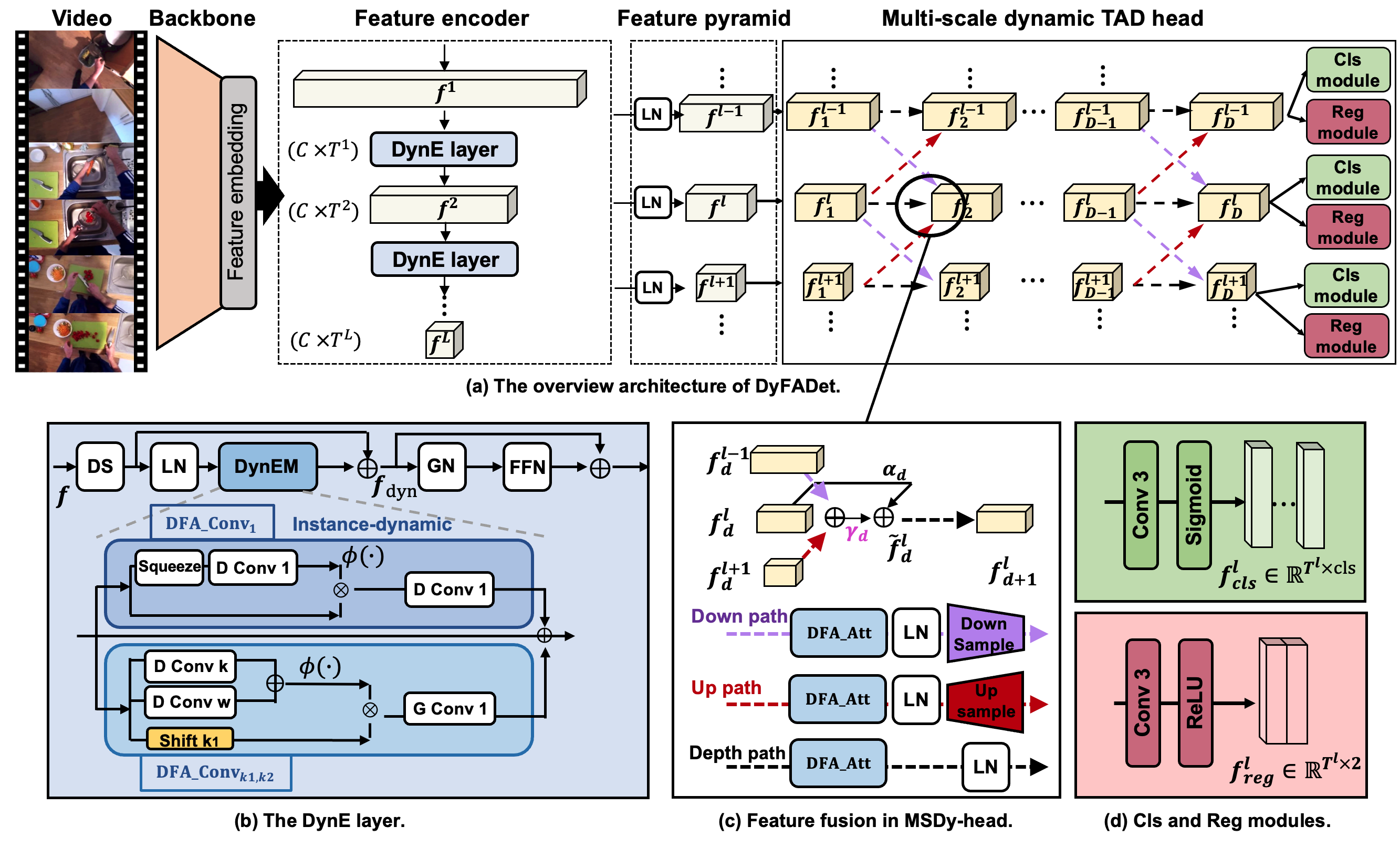}
    \caption{(a) Overview of \nameshort{}. (b) The DynE layer consisting of the feature encoder. GN is Group-normalization~\cite{wu2018group}. (c) The multi-scale feature fusion in \nameshorth{}. (d) The classification and regression module obtains the classification and boundary results.}
    \label{fig_over}
    \vspace{-9pt}
\end{figure*}

As illustrated in Fig.~\ref{fig_over} (b), the \nameshorte{} layer is built by substituting the SA module with the proposed \nameshorte{} module, which follows the marco-architecture of transformer layers~\cite{zhang2022actionformer}. The \nameshorte{} module has two branches: A instance-dynamic branch based on DFA\_Conv with kernel size of 1 generates the global temporal weighted mask to help aggregate global the action information. For another branch, we propose to use the DFA\_Conv with convolutions implemented by different kernel sizes to better generate the weighted mask tenors. To further improve the efficiency during feature learning, all convolution modules are implemented by depth-wise convolutions (D Conv) with the corresponding dimensionality. The two branches can be written as
\begin{small}    
\begin{align}
    \f_{\text{ins}} &= \text{DFA\_Conv}_{1} \left(  \text{Squeeze} ( \text{LN}( \text{DS} (\f)  ) ) \right)  \\
    \f_{\text{k}} &= \text{DFA\_Conv}_{k,w}\left( \text{LN}(\text{DS} (\f)) \right),
\end{align}
\end{small}
where the DS is the down-sampling achieved by max-pooling with the scale of 2. LN is the Layer normalization~\cite{ba2016layer}. The Squeeze is achieved by average pooling at the channel dimension. $w$ is the factor to expand the window size of the convolution to $ w*(k+1)$, which enables the module can learn a the feature with long-term information. Then the output features of the \nameshorte{} module can be represented by 
\begin{small}
\begin{align}
    \f_{\text{dyn}} =\f_{\text{w}} + \f_{\text{ins}} + \text{DS} (\f).
\end{align}
\end{small}

Overall, in the feature encoder, each \nameshorte{} layer will down-sample the feature with a scale of 2 to generate the representations with different temporal resolutions. The dynamic feature selection ability of \nameshorte{} layers will guarantee the discriminative information of the obtained representations and alleviate the less-discriminant feature problem, which leads to better TAD performance. 

\subsubsection{\namelongh{}.} The shared-weight static detection heads in common TAD models can have the head inadaptation issue, which means that the optimal weights for detecting long-range and short-range action instances can be different, resulting in the sub-optimal detection performance. Even when implemented with the cross-level feature fusion, the heads only show limited improvement. Intuitively, fusing the features from a higher scale helps the head to explore more global long-term information. Exploring the information from a lower scale can enable the head to find more boundary details. We infer that such a multi-scale fusion can benefit the detection performance only if these intrinsic representations are properly selected from the adjacent scale. These motivate us to build a novel \nameshortm{}, which can dynamically adjust the shared parameters based on inputs and selectively fuse the cross-level features for better performance.

The architecture of a \nameshorth{} is illustrated in Fig.~\ref{fig_over}~(a). Both the features from the corresponding and the adjacent levels in the pyramid are sent to the \nameshorth{}. By implementing the dynamic feature learning mechanism with the proposed \nameshortm{}, the head parameters can adjust based on inputs, and important representations will be selectively fused. Specifically, suppose that the depth of the head is $D$, the output features $\f_{d+1}^{l}$ at the $l$-th level can be calculated as the accumulation of $\f_{d}^{l-1}$, $\f_{d}^{l}$ and $\f_{d}^{l+1}$, which are processed by three different paths (shown in Fig.~\ref{fig_over}~(c)). The down path and up path are built based on the \nameshortm{} with an additional LN and down-sampling (DS) or up-sampling (US) interpolation module with a scale of 2. The features will be first fused by
\begin{small}
    
\begin{align}
    \Tilde{\f}_{d}^{l}\  =\   &\gamma_{d} \cdot \bigg( \text{DS}(\ \text{DFA\_Att}(\f_{d}^{l-1})\ ) \  +  \nonumber \\
    & \text{US}(\ \text{DFA\_Att}(\f_{d}^{l+1})\ ) \bigg) + \alpha_{d} \cdot \f_{d}^{l},
\end{align}
\end{small}
where $\gamma_{d}$ and $\alpha_{d}$ are two learnable factors. Then the resultant feature will be processed by the depth path by
\begin{small}
    \begin{align}
    \f_{d+1}^{l}\ = \ \text{DFA\_Att}(\ \Tilde{\f}_{d}^{l}\ ).
\end{align}
\end{small}

The proposed \nameshorth{} With the multi-scale dynamic feature fusion procedure, the final features at depth $D$, namely, $\f_{D}^{l}$, $l=1,...,L$, will be used to detect the action instances with different temporal ranges. 


\subsubsection{Classification and regression modules.} The final classification (Cls) and regression (Reg) modules are designed to process final features across all levels. The cls module is realized by using a 1D convolution with a Sigmoid function to predict the probability of action categories at each time stamp. The Reg module is implemented using a 1D convolution with a ReLU to estimate the distances $\{ d_m^s, d_m^e\}$ from the timestamp $t$ to the action start and end $\{ t_m^s, t_m^e\}$ for $m$-th action instance, which can be obtained by $t_m^s = t-d_m^s$ and $t_m^e = t+d_m^e$.

%



\subsection{Training and Inference}


\subsubsection{Training.} Follow~\cite{zhang2022actionformer}, the center sampling strategy~\cite{zhang2022actionformer} is used during training, which means that the instants around the center of an action instance are labeled as positive during training. The loss function of the proposed \nameshort{} follow the simple design in~\cite{zhang2022actionformer}, which has two different terms: (1) $\mathcal{L}_{\text{cls}}$ is a focal loss~\cite{lin2017focal} for classification; (2) $\mathcal{L}_{\text{reg}}$ is a DIoU loss~\cite{zheng2020distance} for distance regression. The loss function is then defined as
\begin{align}
\mathcal{L} =  \sum_{t,l} \big(\mathcal{L}_{cls} + \lambda_{\text{reg}} \mathbb{I}_{ct} \ \mathcal{L}_{reg}\big) / T_{pos}, 
\end{align}
where $T_{pos}$ is the number of positive timestamps and $\mathbb{I}_{ct}$ is an indicator function that denotes if a time step $t$ is within the sampling center of an action instance.

\subsubsection{Inference.} During inference, if a classification score is higher than the detection threshold, the instant will be kept. Then, Soft-NMS~\cite{bodla2017soft} will be further applied to remove the repeated detected instances.

%% file: sec/4_exp.tex
\section{Experiments}

\subsection{Experimental settings}

\subsubsection{Datasets.}
Six TAD datasets, including HACS-Segment~\cite{zhao2019hacs}, THUMOS14~\cite{THUMOS14}, ActivityNet-1.3~\cite{caba2015activitynet}, Epic-Kitchen 100~\cite{damen2022rescaling}, Ego4D-Moment Queries v1.0 (Ego4D-MQ1.0)~\cite{grauman2022ego4d} and FineAction~\cite{liu2022fineaction}, are used in our experiments. ActivityNet-1.3 and HACS are two large-scale datasets with 200 classes of action, containing 10,024 and 37,613 videos for training, 4,926 and 5,981 videos for tests. THUMOS14 consists of 20 sport action classes and contains 200 and 213 untrimmed videos with 3,007 and 3,358 action instances on the training and test set, respectively. The Epic-Kitchen 100 and Ego4D-MQ1.0 are two datasets in first-person vision. Epic-Kitchen 100 has two sub-tasks: noun and verb localization, containing 495 and 138 videos with 67,217 and 9,668 action instances for training and testing, respectively. Ego4D-MQ1.0 has 2,488 video clips and 22.2K action instances from 110 pre-defined action categories, which are densely labeled. FineAction contains 103K temporal instances of 106 fine-grained action categories, annotated in 17K untrimmed videos.

\vspace{-10pt}
\subsubsection{Evaluation metric and experimental implementation.} The standard mean average precision (mAP) at different temporal intersection over union (tIoU) thresholds will be reported as evaluation metric in the experiments. We follow the practice in~\cite{shi2023tridet,zhang2022actionformer} that uses off-the-shelf pre-extracted features as input. Our method is trained with AdamW~\cite{loshchilov2017decoupled} with warming-up. More training details are provided in the supplementary materials.

\vspace{-10pt}
\subsubsection{Detailed architecture of \nameshort{}.} We used 2 convolutions for feature embedding, 7 DynE layers as the encoder, and separate \nameshorth{}s for classification and regression as the detectors. In our experiments, we report the best results of \nameshort{} with different architectures. A comprehensive ablation study about the architecture is also provided in Section~\ref{ab}.

\vspace{-10pt}
\subsection{Main results}

\begin{wraptable}{r}{6.7cm}\footnotesize
    \vspace{-30pt}
    \caption{Results on HACS-segment.}
\vspace{-10pt}
\label{tab:HACS}
\begin{center}
\resizebox{0.99\linewidth}{!}{
\begin{tabular}{l|l|cccc}
\toprule
Method & Backbone & 0.5 & 0.75 & 0.95 & Avg. \\
\midrule
SSN~\cite{zhao2017temporal} & I3D & 28.8 & 18.8 & 5.3 & 19.0 \\
LoFi~\cite{xu2021low} & TSM & 37.8 & 24.4 & 7.3 & 24.6 \\
G-TAD~\cite{xu2020g} & I3D & 41.1 & 27.6 & 8.3 & 27.5 \\
TadTR~\cite{liu2022end} & I3D & 47.1 & 32.1 & 10.9 & 32.1 \\
BMN~\cite{lin2019bmn} & SlowFast & 52.5 & 36.4 & 10.4 & 35.8 \\
ActionFormer~\cite{zhang2022actionformer} & SlowFast & 54.9 & 36.9 & 9.5 & 36.4 \\
TALLFormer~\cite{cheng2022tallformer} & Swin & 55.0 & 36.1 & 11.8 & 36.5 \\
TCANet~\cite{qing2021temporal} & SlowFast & 54.1 & 37.2 & 11.3 & 36.8 \\
TriDet~\cite{shi2023tridet} & SlowFast & 56.7 & 39.3 & 11.7 & 38.6 \\
TriDet~\cite{shi2023tridet} & VM2-g & 62.4 &44.1 &13.1 &43.1 \\
\midrule
\rowcolor{mygray} Ours & SlowFast & 57.8 & 39.8 & 11.8 & 39.2  \\ 
\rowcolor{mygray} Ours & VM2-g  &  \textbf{64.0}  & \textbf{44.8}  & \textbf{14.1}  &\textbf{44.3}  \\
\bottomrule
\end{tabular}}
\end{center}
    \vspace{-30pt}
\end{wraptable}
\subsubsection{HACS.} The performance of different TAD methods on the HACS dataset is provided in Table~\ref{tab:HACS}, where average mAP in [0.5:0.05:0.95] is reported, and the best and the second best performance are denoted by bold and blue. In our experiments, the SlowFast~\cite{feichtenhofer2019slowfast} features are used for the proposed \nameshort{} in TAD tasks on HACS. 

From the results, we see that our method with SlowFast features outperforms all other evaluated methods in terms of average mAP (39.2\%), and also achieves the best performance across all tIoUs thresholds. Notably, with tIoU = 0.5, the \nameshort{} surpasses the Tridet by 1.1\%. As a TAD model can generally benefit from a more advanced backbone, we further implement VideoMAE V2-Gaint~\cite{wang2023videomaev2} (VM2-g) to conduct TAD on HACS. Remarkably, our method achieves the highest performance with VideoMAE V2-Gaint and beats the previous SOTA TriDet (VM2-g) with the new SOTA result on HACS, 44.3\%.

\begin{table*}[!t]
\caption{Comparison with the SOTA methods on THUMOS14 and ActivityNet-1.3. TSN~\cite{wang2016temporal}, I3D~\cite{carreira2017quo}, Swin Transformer (Swin)~\cite{cheng2022tallformer} and TSP (R(2+1)D)~\cite{alwassel2021tsp} features are used. The best and the second best performance are denoted by \textbf{bold} and \textcolor{blue}{blue}.}
\label{tab:THUMOS14}
\vspace{-10pt}
\begin{center}
\resizebox{0.99\linewidth}{!}{
\begin{tabular}{l||l|cccccc||l|cccc}
\toprule
\multirow{2}{*}{Method} &  \multicolumn{7}{|c||}{THUMOS14} & \multicolumn{5}{|c}{ActivityNet-1.3 }\\
\cmidrule{2-8} \cmidrule{9-13} 
 & Features &  0.3 & 0.4 & 0.5 & 0.6 & 0.7 & Avg. & Features &  0.5 & 0.75 & 0.95 & Avg.  \\
\midrule
BMN~\cite{lin2019bmn} & TSN & 56.0 & 47.4 & 38.8 & 29.7 & 20.5 & 38.5 &TSN&50.1&34.8&8.3&33.9\\
G-TAD~\cite{xu2020g}& TSN & 54.5 & 47.6 & 40.3 & 30.8 & 23.4 & 39.3 &TSN&50.4&34.6&9.0&34.1\\
A2Net~\cite{yang2020revisiting} & I3D & 58.6 & 54.1 & 45.5 & 32.5 & 17.2 & 41.6 &I3D&43.6 & 28.7 & 3.7 &  27.8 \\
PBRNet~\cite{liu2020progressive} & I3D & 58.5 &54.6 &51.3 &41.8 &29.5 & - & I3D & 54.0& 35.0& 9.0 &35.0 \\
TCANet~\cite{qing2021temporal}& TSN & 60.6 & 53.2 & 44.6 & 36.8 & 26.7 & 44.3 &TSN&52.3&36.7&6.9&35.5\\
RTD-Net~\cite{tan2021relaxed}& I3D & 68.3 & 62.3 & 51.9 & 38.8 & 23.7 & 49.0 & I3D & 47.2 & 30.7 & 8.6 & 30.8 \\
VSGN~\cite{zhao2021video}& TSN & 66.7 & 60.4 & 52.4 & 41.0 & 30.4 & 50.2 &I3D&52.3&35.2&8.3&34.7\\
AFSD~\cite{lin2021learning}& I3D & 67.3 & 62.4 & 55.5 & 43.7 & 31.1 & 52.0 &I3D&52.4&35.2&6.5&34.3\\
ReAct~\cite{shi2022react}& TSN & 69.2 & 65.0 & 57.1 & 47.8 & 35.6 & 55.0 &TSN&49.6&33.0&8.6&32.6\\
TadTR~\cite{liu2022end}& I3D & 74.8 & 69.1 & 60.1 & 46.6 & 32.8 & 56.7 &TSN&51.3&35.0&9.5&34.6 \\
TALLFormer~\cite{cheng2022tallformer}&Swin & 76.0 & - & 63.2 & - & 34.5 & 59.2 &Swin&54.1&36.2&7.9&35.6\\
ActionFormer~\cite{zhang2022actionformer} & I3D& 82.1 & 77.8 & 71.0  & 59.4 &  43.9  & 66.8 & R(2+1)D & \textcolor{blue}{54.7} &37.8&\textbf{8.4}&36.6\\
ASL~\cite{shao2023action}& I3D  & 83.1 & 79.0 & 71.7 & 59.7 & 45.8 & 67.9  & I3D & 54.1 & 37.4 & 8.0 & 36.2\\
TriDet~\cite{shi2023tridet}& I3D  & \textcolor{blue}{83.6} & \textbf{80.1} & \textbf{72.9} & \textbf{62.4} & 47.4 & \textbf{69.3} &R(2+1)D&\textcolor{blue}{54.7} & \textcolor{blue}{38.0} & \textbf{8.4}& \textcolor{blue}{36.8}\\
\midrule
\rowcolor{mygray} Ours & I3D & \textbf{84.0} & \textbf{80.1} & \textcolor{blue}{72.7} & \textcolor{blue}{61.1} & \textbf{47.9} & \textcolor{blue}{69.2}  & R(2+1)D & \textbf{58.1} & \textbf{39.6} & \textbf{8.4} & \textbf{38.5}\\ 
\bottomrule
\end{tabular}}
\vspace{-20pt}
\end{center}
\end{table*}

\vspace{-10pt}
\subsubsection{THUMOS14 and ActivityNet-1.3.} The experimental results which compare the performance of ours and other TAD models are shown in Table~\ref{tab:THUMOS14}. Average mAP in [0.3:0.1:0.7] and [0.5:0.05:0.95] are reported on THUMOS14 and ActivityNet-1.3, respectively. In the experiments, our method conducts the TAD task based on I3D and R(2+1)D features for the THUMOS14 and ActivityNet-1.3 datasets. We see that our method with I3D features achieves the mAP of 69.2\%, which is competitive to TriDet~\cite{shi2023tridet}, while significantly outperforming all other related TAD methods on THUMOS14. On ActivityNet-1.3, the proposed \nameshort{} significantly surpasses the TriDet by about 1.7\%, and achieves the mAP of 38.5\%. The high performance of the proposed \nameshort{} indicates the effectiveness of dynamic feature learning mechanisms in modern TAD methods.
\begin{wraptable}{r}{7cm}\footnotesize
    \vspace{-30pt}
    \caption{Results on THUMOS14 using VM2-g.}
    \label{tab vae}
    \resizebox{0.99\linewidth}{!}{
    \renewcommand{\arraystretch}{1.1}
    \centering
    \begin{tabular}{l|l|cccc}
    \toprule
        Method & Backbone &0.3 & 0.55 & 0.7 & Avg. \\
    \midrule
        ActionFormer~\cite{zhang2022actionformer} & VM2-g& 84.0 & 73.0 & 47.7& 69.6 \\
        TriDet~\cite{shi2023tridet} & VM2-g& 84.8 & 73.3 & 48.8& 70.1 \\ \midrule
        \rowcolor{mygray}  Ours & VM2-g&  84.3 & 73.7 & \textbf{50.2}  & 70.5 \\
        \rowcolor{mygray}  Ours & VM2-g+F & \textbf{85.4} & \textbf{74.0} & \textbf{50.2}  & \textbf{71.1}\\
    \bottomrule
    \end{tabular}}
    \vspace{-20pt}
\end{wraptable}

Generally, a TAD model can benefit from a more advanced backbone. Therefore, we further implement VideoMAE V2-Gaint~\cite{wang2023videomaev2} (VM2-g) to conduct TAD on THUMOS14. We see that all TAD methods achieve significant improvements with an advanced feature extraction backbone, VM2-g. While, our method can achieve the mAP of 70.5\%, which is superior to TriDet and ActionFormer. Also, the detection performance can be boosted to 71.1\% if we additionally use the optic flow (F) features.

\begin{table}[htp]

\setlength{\tabcolsep}{3pt}
  \begin{minipage}[b]{0.5\linewidth}
    \centering
    \caption{Results on Epic-Kitchen~100. }
    \label{tab:Epic-Kitchen}
    \vspace{-10pt}
    \resizebox{1.0\linewidth}{!}{
    \begin{tabular}{l|l|cccccc}
    \toprule
    & Method & 0.1 & 0.2 & 0.3 & 0.4 & 0.5 & Avg. \\
    \midrule
    \multirow{4}{*}{V.}& G-TAD~\cite{xu2020g} & 12.1 & 11.0 & 9.4 & 8.1 & 6.5 & 9.4 \\
    & ActionFormer~\cite{zhang2022actionformer} & 26.6 & 25.4 & 24.2 & 22.3 & 19.1 & 23.5 \\
    & ASL~\cite{shao2023action} & 27.9 & - & 25.5 & - & 19.8 & 24.6 \\
    \cmidrule{2-8}
    \rowcolor{mygray} & Ours & \textbf{28.0} & \textbf{27.0} & \textbf{25.6} & \textbf{23.5} & \textbf{20.8} & \textbf{25.0}\\
    \midrule
    \multirow{4}{*}{N.}& G-TAD~\cite{xu2020g} & 11.0 & 10.0 & 8.6 & 7.0 & 5.4 & 8.4 \\
    & ActionFormer~\cite{zhang2022actionformer} & 25.2 & 24.1 & 22.7 & 20.5 & 17.0 & 21.9 \\
    & ASL~\cite{shao2023action} & 26.0 & -& 23.4 & -& 17.7 & 22.6 \\
    \cmidrule{2-8}
    \rowcolor{mygray} & Ours & \textbf{26.8} & \textbf{26.0} & \textbf{24.1} & \textbf{21.9} & \textbf{18.5} & \textbf{23.4}  \\
    \bottomrule
    \end{tabular}}
  \end{minipage}\hfill
  \begin{minipage}[b]{0.5\linewidth}
    \centering
    \caption{Results on Ego4D-MQ1.0.}
    \vspace{-10pt}
    \label{tab:ego}
    \resizebox{0.98\linewidth}{!}{
    \begin{tabular}{l|l|cccc}
    \toprule
     Method & Features & 0.1 &  0.3 & 0.5 & Avg. \\
    \midrule
    VSGN~\cite{zhao2021video} & SF & 9.1 & 5.8 & 3.4 & 6.0\\
    VSGN~\cite{lin2022egocentric} & EV & 16.6 & 11.5 & 6.6 & 11.4  \\
    ActionFormer~\cite{zhang2022actionformer} & SF & 20.1	& 14.4 & 10.0&	14.9 \\
    ActionFormer~\cite{zhang2022actionformer} & EV & 26.9 & 20.2 & 13.7 & 20.3 \\
    ActionFormer~\cite{zhang2022actionformer} & EV+SF & 28.0 &	21.2	&	15.6	& 21.6 \\
    \midrule
    \rowcolor{mygray} Ours & SF  & 19.0 & 15.0 & 11.2 & 15.3\\ 
    \rowcolor{mygray} Ours & EV & 28.4 & 22.1 & 16.2 & 22.2 \\ 
    \rowcolor{mygray} Ours & EV+SF & \textbf{28.8} & \textbf{22.6} & \textbf{16.9} & \textbf{22.8} \\ 
    \bottomrule
    \end{tabular}}

  \end{minipage}\hfill
  \vspace{-20pt}
\end{table}
\subsubsection{Epic-Kitchen~100 and Ego4D-MQ1.0.} The evaluations are also conducted on two large-scale egocentric datasets, which are shown in Table~\ref{tab:Epic-Kitchen} and Table~\ref{tab:ego}, respectively. For Epic-Kitchen~100, the average mAP in [0.1:0.1:0.5] is reported and all
methods use the SlowFast features. We see that our method has the best performance across all tIoU thresholds on both subsets and achieves an average mAP of 25.0\% and 23.4\% for verb and noun subsets, respectively, which are significantly superior to the strong performance of the recent TAD methods, including Actionformer~\cite{zhang2022actionformer}, ASL~\cite{shao2023action}. For Ego4D-MQ1.0, two types of features, including SlowFast (SF) and EgoVLP (EV) features are used in the experiments. With SlowFast features, the proposed method achieves the mAP of 15.3\%, which significantly outperforms the Actionformer. Moreover, we see that using or combining the features with advanced backbone models, such as EgoVLP~\cite{lin2022egocentric}, can further boost the performance of our method by a large margin.
SF and EV denote Slowfast~\cite{feichtenhofer2019slowfast} and EgoVLP~\cite{lin2022egocentric} features. V. and N. denote the \emph{verb} and \emph{noun} sub-tasks.

\begin{wraptable}{r}{6.5cm}\footnotesize
    \vspace{-10pt}
    \caption{Results on FineAction.}
    \label{tab fa}
    \resizebox{0.99\linewidth}{!}{
    \renewcommand{\arraystretch}{1.1}
    \centering
    \begin{tabular}{l|l|cccc}
    \toprule
        Method & Backbone &0.5 & 0.75 & 0.95 & Avg. \\
    \midrule
        BMN & I3D & 	14.4	& 8.9&	3.1& 9.3 \\
        G-TAD & I3D & 	13.7	& 8.8&	3.1& 9.1 \\
        ActionFormer& InternVideo& - & -& -& 17.6 \\
        ActionFormer & VM2-g& 29.1	& 17.7	& 5.1 & 18.2\\ \midrule
        \rowcolor{mygray} Ours & VM2-g & \textbf{37.1} & \textbf{23.7} & \textbf{5.9} & \textbf{23.8}\\
    \bottomrule
    \end{tabular}}
    \vspace{-20pt}
\end{wraptable}
\subsubsection{FineAction.} In the experiments, we report the performance of the different popular TAD methods including BMN~\cite{lin2019bmn}, G-TAD~\cite{xu2020g}, ActionFormer~\cite{zhang2022actionformer}, and our method on TAD task. Moreover, I3D~\cite{wang2016temporal}, InternVideo~\cite{wang2022internvideo} and and VM2-g~\cite{wang2023videomaev2} are used to extract the off-line features and the average mAP in [0.50:0.05:0.95] is reported for all methods. The experimental results are provided in Tab.\ref{tab fa}. The experimental results show that our method outperforms other TAD models and achieve a new SOTA result on FineAction, 23.8\%, with the VideoMAEv2-giant.

\subsection{Ablation study}
\label{ab}
Ablation studies are conducted on THUMOS14 to explore more properties about the \nameshortm{} and \nameshort{}.

\begin{wraptable}{r}{6cm}\footnotesize
    \vspace{-25pt}
    \caption{Results with different modules.}
    \label{tab ab}
    \resizebox{\linewidth}{!}{
    \begin{tabular}{l|ll|c}
    \toprule
     Method & Encoder & MS-head  &  Avg. \\
    \midrule
    Baseline & Conv &  \faTimes  & 62.1 \\
    Baseline & DeformConv~\cite{lei2018temporal} & \faTimes&  66.1 \\
    Baseline & SE~\cite{hu2018squeeze} & \faTimes&  63.4 \\
    Baseline & Dyn Conv~\cite{zhou2023interpret} & \faTimes&  66.7 \\
    Baseline & TD3d Conv~\cite{li2020spatio} & \faTimes&  66.5 \\
    \nameshort{}$^*$   & DFA\_Conv & \faTimes& 66.8 \\
    ActionFormer & SA & \faTimes &  66.8\\    
    \nameshort{}$^\dag$ & DynE & \faTimes &  67.8  \\
    \midrule
    \nameshort{}$^\ddag$ & Conv & Dyn &  67.9 \\
    \nameshort{}$^\ddag$ & DynE & Conv &  68.0 \\
    \nameshort{} & DynE & Dyn & \textbf{69.2} \\
    \bottomrule
    \end{tabular}}
    \vspace{-10pt}
\end{wraptable}
\subsubsection{\nameshortm{} modules in TAD.} The experiments investigate the effectiveness of different dynamic feature aggregation modules in TAD tasks. The results are shown in Table~\ref{tab ab}, where different implementations of the encoder and the detection head are evaluated. The baseline model is realized by an all-convolution TAD model, which achieved the mAP of 62.1\%. We further use different dynamic modules to substitute or improve the convolutions in the encoder as the comparison, such as the deformable 1-d convolution~\cite{lei2018temporal}, squeeze-and-excitation module~\cite{hu2018squeeze}, Dynamic convolution~\cite{zhou2023interpret} and temporal deformable 3d convolution module~\cite{li2020spatio}. All dynamic modules achieve better performance than the baseline model with convolution, indicating the strong ability of dynamic modules in TAD tasks. While, due to the stronger adaptation ability of \nameshortm{}, the \nameshort{}$^*$ substituting the convolutions with \nameshortm{} can achieve the performance equaling to the recent strong TAD model, Actionformer. Moreover, using the proposed \nameshorte{} layer (\nameshort{}$^\dag$) further increases the final performance by 1.0\%. For the detection head, we see that applying the multi-scale connection in the TAD head can improve the final detection performance. However, naively using the convolution to connect different scales (\nameshort{}$^\ddag$) only results in limited improvements. While, after being equipped with \nameshorth{}, the \nameshort{} can achieve a performance of 69.2\%, outperforming all other models in the experiments. A more comprehensive study about the architecture for \nameshort{} is provided in supplementary materials.

\vspace{-5pt}
\subsubsection{The discriminability of the learned features.} 
\begin{figure}[!t]
	\centering
        \includegraphics[width=1.0\linewidth]{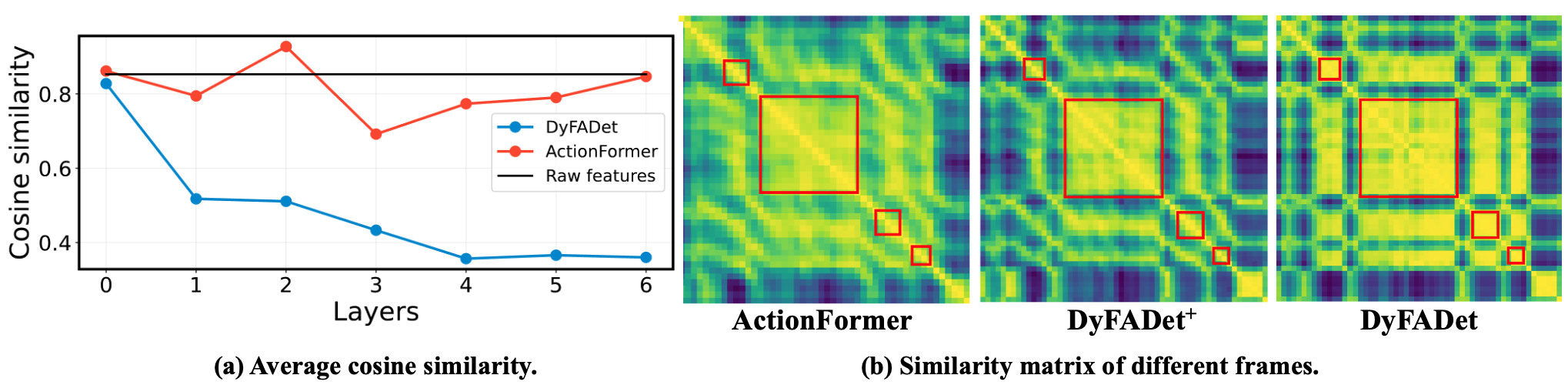}
	\caption{(a) The average cosine similarity between features at different timestamps in the same level among each encoder layer. (b) Similarity matrix of the extracted features among timestamps.}
	\label{fig_sim}
\end{figure}

As shown in~\cite{shi2023tridet,tang2023temporalmaxer}, the features obtained by recent TAD methods~\cite{zhang2022actionformer,weng2022efficient} tend to exhibit high similarities between snippets, which leads to the less-discriminant feature problem and be harmful to TAD performance. In Fig.~\ref{fig_sim}(a), we perform statistics of the average cosine similarity between each feature at different timestamps. We observe that the features in Actionformer exhibit high similarity, indicating poor discriminability. Using the \nameshorte{} based encoder can address the issue and therefore improve the detection performance. Moreover, we further provide the feature similarity matrix between the different timestamps in Fig.\ref{fig_sim}(b), where the red boxes exhibit the action intervals. The darker color means the features are more discriminant and share less similarity. From the result, we see that the inter-class features from our method within the red boxes show strong similarities, resulting in that the boundary features are distinctive and can be easily extracted. While Actionformer fails to explore the discriminant features during feature learning. From the results, our method addresses the mentioned first issue in Fig.\ref{fig_1} based on the more discriminant learned features in \nameshort{}$^\dag$, which can be further enhanced by using the proposed \nameshort{} as TAD model.

\vspace{-10pt}
\subsubsection{Visualization of \nameshortm{} in \nameshorth{}.}
\begin{figure}[t!]
    \centering
    \includegraphics[width=1.0\linewidth]{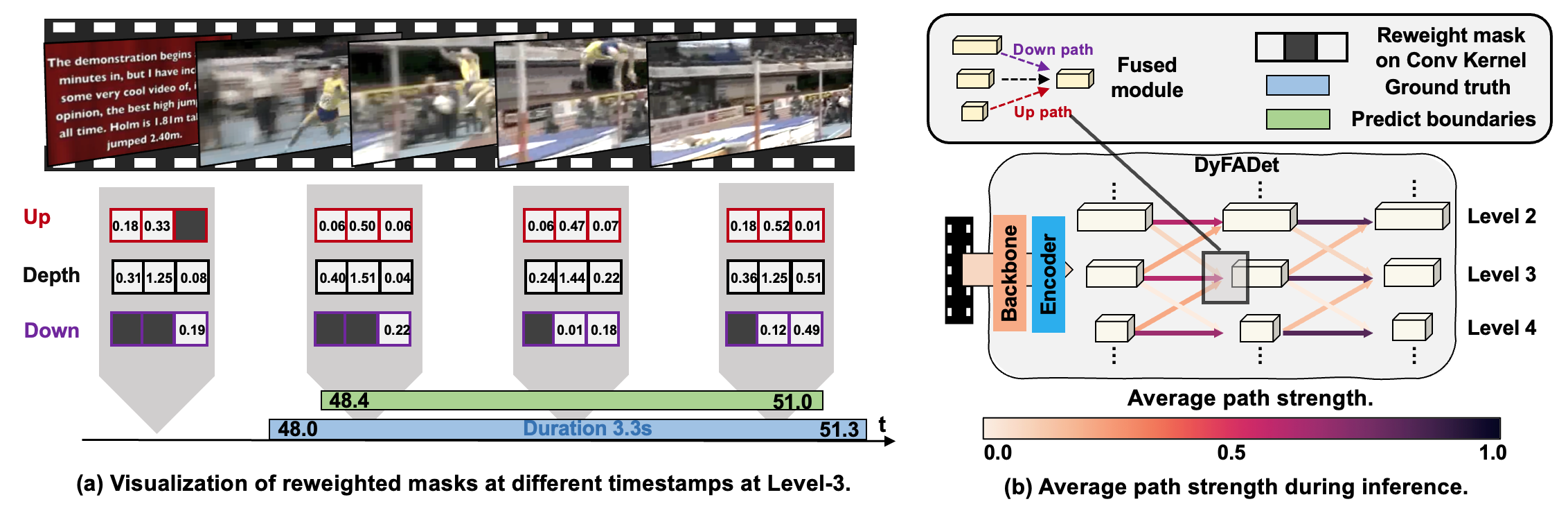}
    \vspace{-10pt}
    \caption{Visualization when \nameshort{} detecting action High Jump. (a) the \nameshortm{} reweight masks at different timestamps at Level-3 in \nameshorth{}. (b) Average path strength during inference of the whole video.}
    \label{fig_vis}
\end{figure}

In Fig.~\ref{fig_vis} (a), we provide the visualization results of the proposed \nameshortm{}. For better visualization, we use the \nameshort{} with $C_m=k$ in Eq.(\ref{eq2}), meaning that the same receptive fields will be shared among channels while varying at different timestamps. From the results we see that the \nameshortm{} can adapt the re-weight masks based on inputs, leading to different formations at different timestamps. Noting that in classical TAD models, such as Actionformer~\cite{zhang2022actionformer}, the parameters of the detection head are shared among different levels, which might be harmful in detection. While, the dynamic properties and multi-scale fusion of the \nameshort{} specialize the detection head for inputs and target action instances, leading to a fine-grained dynamic action detection manner which can achieve better TAD performance. 

Moreover, we visualize the average activated rate at each path for the whole video in \nameshorth{} (shown in Fig.~\ref{fig_vis} (b)). The results show that the proposed \nameshorth{} achieves the dynamic routing-like feature learning mechanism similar to~\cite{song2020fine}. However, our method can simultaneously adapt the receptive fields and the kernel weights, which improves the model ability in TAD tasks. More visualization results can be found in supplementary materials.

\begin{table}[htp]
\setlength{\tabcolsep}{3pt}
  \begin{minipage}[b]{0.44\linewidth}
    \centering
    \caption{Different hyper-parameters. }
    \label{tab:hyper}
    \vspace{-4pt}
    \resizebox{0.8\linewidth}{!}{
    \begin{tabular}{cc|cc|cc}
    \toprule
    $w$ &mAP & D & mAP & DynType & mAP\\
    \midrule
    3 &  68.2 & 2 & 67.4 & C  & 68.4 \\
    5 &  69.2 & 3 & 69.2 & K  & 69.2 \\
    6 &  67.0 & 4 & 67.1 & CK & 68.4 \\
    \bottomrule
    \end{tabular}}
  \end{minipage}\hfill
  \begin{minipage}[b]{0.55\linewidth}
    \centering
    \caption{Average latency on THUMOS14.}
    \vspace{-10pt}
    \label{tab:dy-speed}
    \resizebox{0.98\linewidth}{!}{
    \begin{tabular}{l|cc|cc|c}
\toprule
\multirow{2}{*}{Method}& \multicolumn{2}{c}{AP}  & \multicolumn{2}{c}{Latency} & Params \\
  & 0.5 &Avg. &  CPU(s) & GPU(ms)  &  (M)\\
\midrule
ActionFormer &  71.0 & 66.8 & 0.57 & 68.4 & 29.2 \\
\nameshort{}$^\dag$& 71.7 &  67.8 & \textbf{0.16} & \textbf{44.7} & 12.7 \\
\nameshort{} & \textbf{72.7} & \textbf{69.2}  & 0.49 &58.7 & 18.8 \\
\bottomrule
\end{tabular}}

  \end{minipage}\hfill
  \vspace{-10pt}
\end{table}

\subsubsection{Hyper-parameters.} In Table~\ref{tab:hyper} , we further evaluate 
the performance of \nameshort{} on THUMOS14 with different hyper-parameters, including the expanded factor, $w$, the layer number of the detection head, and the dynamic type of the proposed module. We observe the model works best with $D=3$ and the w should be selected for different datasets. Moreover, although the dynamic type can affect the final performance, \nameshort{} with each dynamic type achieves higher performance compared to most TAD methods in Table~\ref{tab:THUMOS14}.
\vspace{-10pt}

\subsubsection{Latency.} We test the average latency for single video inference on GeForce RTX 4090 GPU on THUMOS14. As shown in Table~\ref{tab:dy-speed}, \nameshort{}$^\dag$ is faster than ActionFormer while has better detection performance. Moreover, \nameshort{} can be improved by the \nameshorth{} which further brings 1.4\% average mAP improvement and the latency is still comparable to Actionformer. 
\vspace{-10pt}

%% file: sec/5_conclusion.tex
\section{Conclusion}
\label{sec:conclusion}
In this paper, we introduced a novel \nameshortm{} module simultaneously adapting its kernel weights and receptive fields, to address the less-discriminant feature and head inadaptation issues in TAD models. The proposed \nameshort{} based on \nameshortm{} achieves high performance on a series of TAD benchmarks, which indicates that an input-based fine-grained feature extraction mechanism should be considered for building high-performance TAD models. For future works, we believe that the efficiency of \nameshort{} can be further improved by combining the sparse convolution~\cite{spconv2022} and adding additional constraints to encourage each \nameshortm{} to mask as many features as possible with a minor performance penalty. The applications of \nameshortm{} in more video-understanding tasks will be further investigated. 

\subsubsection{Acknowledgement.} This work is supported in part by National Natural Science Foundation of China under Grants 62206215, China Postdoctoral Science Foundation under Grants 2022M712537, and China National Postdoctoral Program for Innovative Talents BX2021241.



%% file: sec/sup1.tex
\newpage

\section{Implementation Details}

We now present implementation details including the network architecture, training and inference in our experiments. Further details can be found in our code.

\subsection{Network architecture} 

In Fig.~\ref{figs_over}, we present our network architecture. Specifically, videos will be first processed by a given backbone model to extract the features of the videos. Then these pre-extracted features will be used as the inputs of the TAD model. Feature feature embedding layers based on 2 layers of convolutions followed by LN~\cite{ba2016layer} will first used to calculate the input video features. After that, a series of DynE layers are implemented as the encoder, where the first two DynE layers are serviced as the stem and the rest will successfully downsample the temporal resolution of the features will a scale of 2. 

Following the common settings in~\cite{zhang2022actionformer,shi2022react,shi2023tridet}, a ``2+5'' encoder architecture will be used for each dataset except for Ego4D MQ, meaning that 2 stem layers and 5 downsampling layers will be used for building the encoder, and the outputs of the last 5 layers will be used to build the feature pyramid. Ego4D MQ applies a ``2+7'' encoder architecture following the design in~\cite{zhang2022actionformer}. 

The outputs from the downsampling layers will be first processed by LNs~\cite{ba2016layer} and then used to build the feature pyramid. Then we use separate MSDy-heads for classification and regression as the detectors. In our experiments, the 2 adjacent-level features will be sent to the detection head during detection. For instance, if we detect the action instances at the 4-th level, then the features from the 3-rd and 5-th levels will be also sent to the heads. Specially, for the last level, only features from the previous level will be used as inputs. The \nameshorth{} will share the parameters while detecting the action instances at different feature levels.

\subsection{Training details} 

During training, we randomly selected a subset of consecutive clips from an input video and capped the input length to 2304, 768, 960, 2304, and 1024 for THUMOS14, ActivityNet-1.3, HACS, Epic Kitchens 100, and Ego4D MQV1.0, respectively. Model EMA and gradient clipping are also implemented to further stabilize the training. We follow the practice in~\cite{shi2023tridet,zhang2022actionformer} that uses off-the-shelf pre-extracted features as input. Our method is trained with AdamW~\cite{loshchilov2017decoupled} with warming-up and the learning rate is updated with Cosine Annealing schedule~\cite{loshchilov2016sgdr}. Moreover, hyper-parameters are slightly different across datasets and discussed later in our experiment details. More details can be found in our code. We now describe our experiment details for each dataset:

\begin{figure*}[t!]
    \centering
    \includegraphics[width=0.95\textwidth]{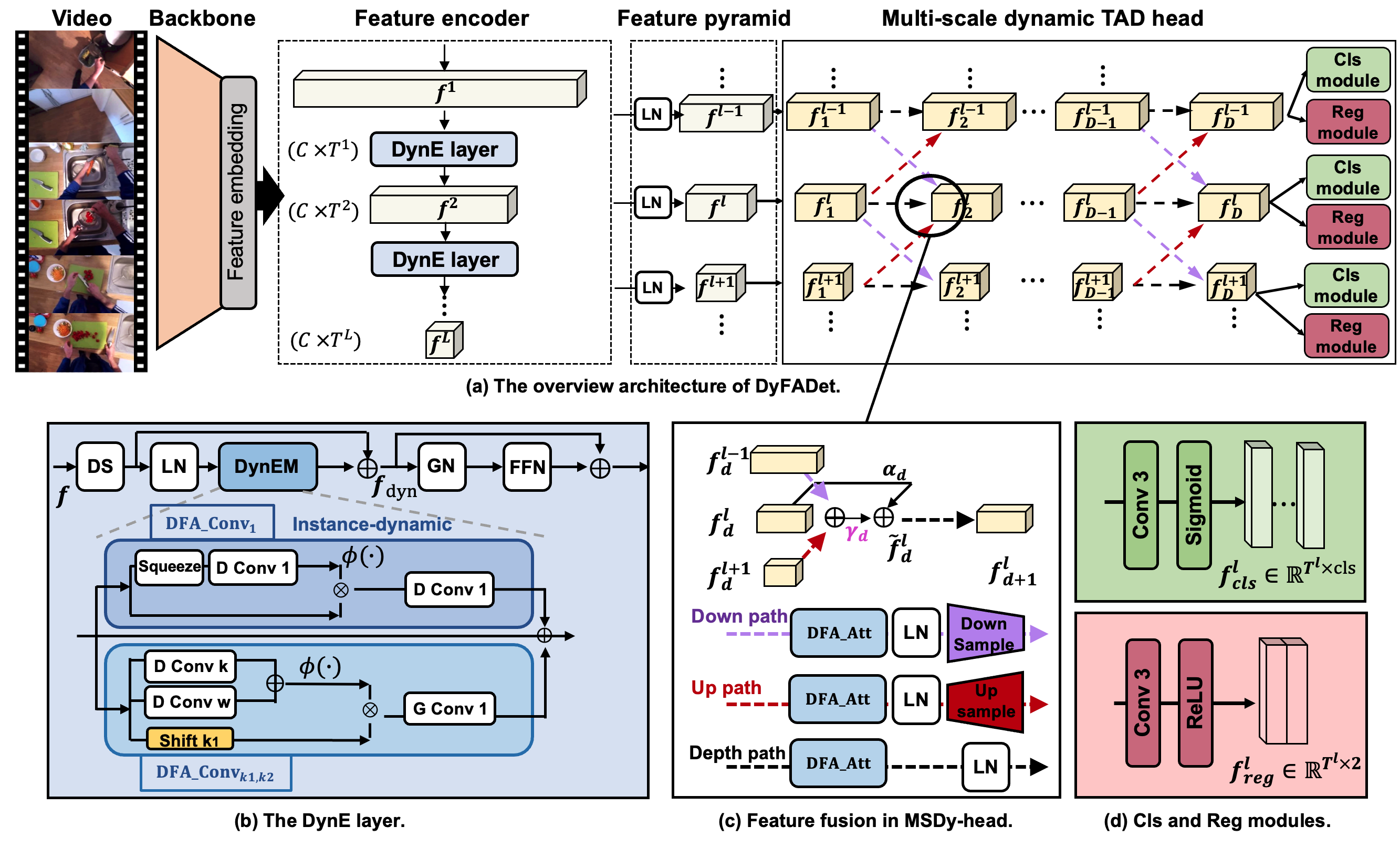}
    \caption{(a) Overview of \nameshort{}. (b) The DynE layer consisting of the feature encoder. GN is Group-normalization~\cite{wu2018group}. (c) The multi-scale feature fusion in \nameshorth{}. (d) The classification and regression module obtains the classification and boundary results.}
    \label{figs_over}
\end{figure*}

\begin{itemize}
    \item \textbf{THUMOS14:} We used two-stream I3D~\cite{carreira2017quo} pretrained on Kinetics to extract the video features on THUMOS14. VideoMAE V2~\cite{wang2023videomaev2} is further implemented to improve the performance of our method. Following~\cite{shi2023tridet}, the initial learning rate is set to $1e-4$ with a batch size of 2. We train 40 epochs for THUMOS14 containing warmup 20 epochs with a weight decay of $2.5e-2$.
    
    \item \textbf{ActivityNet-1.3} We used TSP features~\cite{alwassel2021tsp} pretrained on Kinetics to extract the video features. Following~\cite{shi2023tridet}, the initial learning rate is set to $5e-4$. We train 15 epochs for ActivityNet-1.3 containing warmup 5 epochs with a weight decay of $5e-2$.
    
    \item \textbf{HACS} We used SlowFast features~\cite{feichtenhofer2019slowfast} pretrained on Kinetics to extract the video features on HACS. VideoMAE V2~\cite{wang2023videomaev2} is also implemented to test the performance of our method. Following~\cite{shi2023tridet}, the initial learning rate is set to $5e-4$ with a batch size of 8. We train 14 epochs containing warmup 7 epochs with a weight decay of $2.5e-2$.

    \item \textbf{Epic-Kitchen 100} We used SlowFast features~\cite{feichtenhofer2019slowfast} for our method in experiments. For both subsets, we train 30 epochs containing warmup 15 epochs with a weight decay of $5e-2$. and the initial learning rate is set to $2e-4$ with a batch size of 2.

    \item \textbf{Ego4D MQv1.0} We used SlowFast features~\cite{feichtenhofer2019slowfast} and EgoVLP features~\cite{lin2022egocentric} in the experiments. For all settings, we train 15 epochs containing warmup 5 epochs with a weight decay of $5e-2$. and the initial learning rate is set to $2e-4$ with a batch size of 2.

    \item \textbf{FineAction} We used VideoMAE V2~\cite{wang2023videomaev2} as the feature extractor for our method. In the experiments, the initial learning rate is set to $5e-4$ with a batch size of 8. We train 14 epochs containing warmup 7 epochs with a weight decay of $2.5e-2$.
    
\end{itemize}

\subsection{Inference details} 

During inference, we fed the full sequence into our model. If a classification score is higher than the detection threshold, the instant will be kept. Then, Soft-NMS~\cite{bodla2017soft} will be further applied to remove the repeated detected instances. For our experiments on ActivityNet-1.3 and HACS, we consider score fusion using external classification scores following the settings in~\cite{zhang2022actionformer,shi2023tridet}. Specifically, given an input video, the top-2 video-level classes given by external classification scores were assigned to all detected action instances in this video, where the action scores from our model were multiplied by the external classification scores. Each detected action instance from our model thus creates two action instances. More details can be found in our code.

\section{Ablation study of the architecture design}
\label{sab}

All experiments in this section are conducted on THUMOS14 to explore more properties about the \nameshortm{} and \nameshort{}.

We provide more comprehensive experimental results to show how the architecture design of the \nameshort{} can affect its TAD performance. The experiments investigate the effectiveness of different dynamic feature aggregation modules in TAD tasks. The results are shown in Table~\ref{tab:dy-ab}, where different implementations of the encoder and the detection head are evaluated. 

The results in the upper panel of the table are used to evaluate the TAD performance without multi-scale connections in detection heads. The baseline model is realized by an all-convolution TAD model, which achieved the mAP of 62.1\%. We further use the deformable 1-d convolution~\cite{dtcn23} to substitute the convolutions in the encoder as the comparison achieves the mAP of 66.1\%. The superiority of the deformable-based model demonstrates the strong ability of dynamic modules in TAD tasks. While, due to the stronger adaptation ability of \nameshortm{}, the \nameshort{}$^*$ substituting the convolutions with \nameshortm{} can achieve the performance equaling to the recent strong TAD model, Actionformer. Moreover, using the proposed \nameshorte{} layer (\nameshort{}$^\dag$) further increases the final performance by 1.0\%. 

Moreover, in the middle panel, we see that applying the multi-scale connection in the TAD head can improve the final detection performance. However, naively using the convolution to connect different scales (\nameshort{}$^\ddag$) only results in limited improvements. While, after being equipped with \nameshorth{}, the \nameshort{} can achieve a performance of 69.2\%, outperforming all other models in the experiments. We also use the basic Conv encoder that is built based on all convolution layers attaching with the proposed \nameshorth{}, which achieves the detection performance of 65.8\%, outperforming the baseline model by 3.7\%. Such a result further demonstrates the effectiveness of the proposed \nameshorth{} in TAD tasks.

In the bottom panel, we test our proposed \nameshort{} with different dynamic properties by controlling the $C_m$ as we describe in Section 3 of the main paper. In our experiments, the $C_m$ is set as the values of $k$, $Ck$, and $C$ (represented by k, ck, and c in Table~\ref{tab:dy-ab}), where $k$ is the kernel size and the $C$ is the number of input channels. From the results, we see that the \nameshort (c) has the best performance on THUMOS14. While other variants of \nameshort can still achieve competitive detection performance compared to the evaluated TAD models in our main paper.

\begin{table}[t]
\caption{Ablation study for the choices of main components in TAD models. MS-head means if using the multi-scale connection in the detection head.}
\label{tab:dy-ab}
\vspace{-10pt}
\begin{center}
\resizebox{0.6\linewidth}{!}{
\begin{tabular}{l|cc|ccc}
\toprule
 Method & Encoder & MS-head & 0.3 & 0.7 &  Avg. \\
\midrule
Baseline & Conv &  \faTimes & 76.3 & 40.8 & 62.1 \\

Deformable model & DeformConv & \faTimes& 81.0 &43.4 & 66.1 \\
\nameshort{}$^*$   & DFA\_Conv & \faTimes&  81.6& 44.6 &66.8 \\

ActionFormer & SA & \faTimes &  82.1 & 43.9 & 66.8\\
\nameshort{}$^\dag$ & DynE & \faTimes & 83.4 & 44.8 & 67.8  \\

\midrule
Conv Encoder & Conv & Dyn & 80.4 & 44.2 & 65.8 \\

\nameshort{}$^\ddag$ & DynE & Conv & 83.1 & 46.4 & 68.0 \\

\nameshort{} & DynE & Dyn & \textbf{84.0} & \textbf{47.9} & \textbf{69.2} \\
\midrule

\nameshort{}(k) & DynE & Dyn & 82.4 & 46.3 & 68.4 \\
\nameshort{}(ck) & DynE & Dyn & 82.7 & 47.1 & 68.4  \\
\nameshort{}(c) & DynE & Dyn & \textbf{84.0} & \textbf{47.9} & \textbf{69.2} \\

\bottomrule
\end{tabular}}
\end{center}
\vspace{-10pt}
\end{table}

\begin{figure*}[htp]
    \centering
    \includegraphics[width=1.0\linewidth]{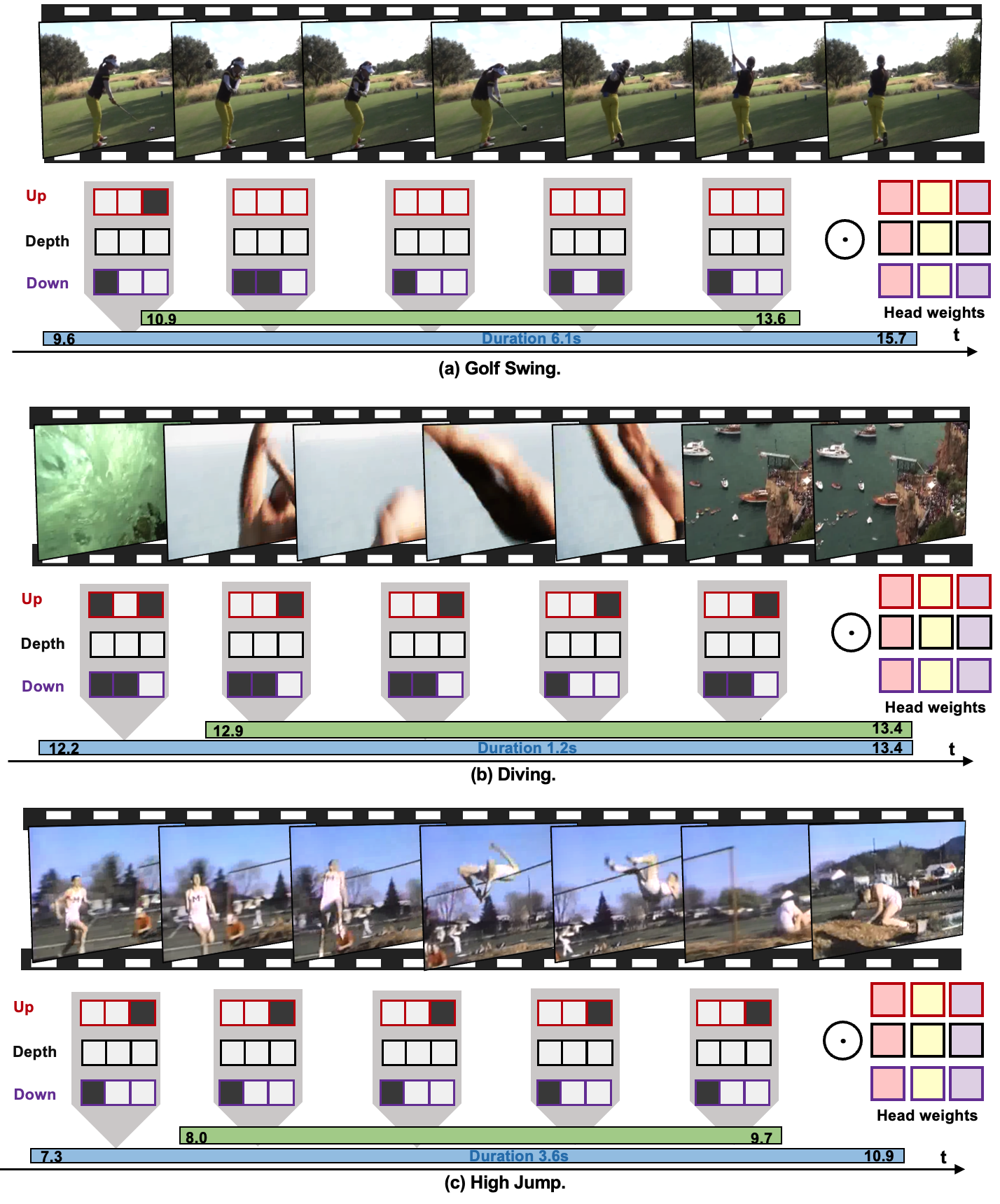}
    \caption{Visualization when \nameshort{} detecting action (a) Golf Swing, (b) Diving and (c) High Jump. The \nameshortm{} reweight masks at different timestamps at the corresponding levels in \nameshorth{} are shown in the figures. Here, we only represent the masked positions of the re-weight masks.}
    \label{figs_vis}
\end{figure*}

\section{Additional Visualizations of \nameshorth{}}

We provide more visualization results of the proposed \nameshortm{} to further show how the proposed dynamic feature learning mechanism can effectively solve the head inadaptation issue in classical TAD models. Following the experiments in the main paper, we use the \nameshort{} with $C_m=k$, meaning that the same receptive fields will be shared among channels while varying at different timestamps. Noting that although the sampling positions are shared among different channels, the kernel weights will still be adjusted based on the inputs. The results are shown in Fig.~\ref{figs_vis}.

From the results we see that the \nameshortm{} can adapt the re-weight masks based on inputs, leading to different formations at different timestamps. Noting that in classical TAD models, such as Actionformer~\cite{zhang2022actionformer}, the parameters of the detection head are shared among different levels, which might be harmful in detection. While, the dynamic properties and multi-scale fusion of the \nameshort{} specialize the detection head for inputs and target action instances, leading to a fine-grained dynamic action detection manner that can achieve better TAD performance. 

The visualization results also show that the weight of the depth path will always be used without masking, indicating the importance of the depth paths. This meets our intuition that the depth path is exclusively designed for action detection. Moreover, we found that the up paths usually play a more important role in action detection, which might be because 1) The high-level features are extracted by more encoder layers in the feature encoder, resulting in more high-level semantic information which benefits the detection. 2) Only a few features from the low-level with high temporal resolution are needed for action detection regardless of the instance duration. 3) The coarse-to-fine feature fusion in \nameshorth{} is similar the feature learning process in~\cite{yang2020resolution}, which demonstrates the importance of the low-frequency information w.r.t temporal dimension. 

We also observe that the later detailed representations generally play a more important role during feature fusion. which might be due to the short-term information from the future can be beneficial for predicting the ending of the action. Moreover, the visualization results show that the predicted action length is generally shorter than the ground truth. We infer that this might be due to that the proposed \nameshort{} only predicts the intrinsic temporal range of the action instance. As shown in in Fig.~\ref{figs_vis} (a), the \nameshort{} thinks the action Golf Swing begins from the time when the person starts to swing the golf club, while ends after the golf club hits the ball. However, the annotation contains the preparation action and the ending action of the person standing after swinging. Intuitively, both of the temporal ranges can be viewed as the action Golf Swing. The examples in Fig.~\ref{fig_vis} (b) and Fig.~\ref{fig_vis} (c) also show the similar trends.